\documentclass{article}

\usepackage{microtype}
\usepackage{graphicx}
\usepackage{subfigure}
\usepackage{booktabs} 

\usepackage{hyperref}

\usepackage[accepted]{icml2025}

\usepackage{amsmath}
\usepackage{amssymb}
\usepackage{mathtools}
\usepackage{amsthm}
\usepackage{graphicx}

\usepackage{listings}

\usepackage{booktabs}
\usepackage{pifont}
\usepackage{threeparttable} 

\usepackage[capitalize,noabbrev]{cleveref}

\theoremstyle{plain}

\theoremstyle{definition}

\theoremstyle{remark}

\usepackage[textsize=tiny]{todonotes}

\usepackage{textcomp}               
\icmlsetsymbol{work}{\textdagger}

\icmltitlerunning{Too Big to Think}

\begin{document}

\twocolumn[
\icmltitle{Too Big to Think: Capacity, Memorization, and Generalization in Pre‑Trained Transformers}




\begin{icmlauthorlist}
\icmlauthor{Joshua Barron}{yyy,work}
\icmlauthor{Devin White}{comp}
\end{icmlauthorlist}

\icmlaffiliation{yyy}{Amazon}
\icmlaffiliation{comp}{Army Educational Outreach Program}
\icmlcorrespondingauthor{Joshua Barron}{josh-ee@berkeley.edu}
\icmlkeywords{Small Language Models, Model Capacity, Pre-training dynamics}

\vskip 0.3in
]



\printAffiliationsAndNotice{\icmlDisclaimer}  

\begin{abstract}
The relationship between memorization and generalization in large language models (LLMs) remains an open area of research, with growing evidence that the two are deeply intertwined. In this work, we investigate this relationship by pre-training a series of capacity-limited Transformer models from scratch on two synthetic character-level tasks designed to separately probe generalization (via arithmetic extrapolation) and memorization (via factual recall). We observe a consistent trade-off: small models extrapolate to unseen arithmetic cases but fail to memorize facts, while larger models memorize but fail to extrapolate. An intermediate-capacity model exhibits a similar shift toward memorization. When trained on both tasks jointly, no model (regardless of size) succeeds at extrapolation. These findings suggest that pre-training may intrinsically favor one learning mode over the other. By isolating these dynamics in a controlled setting, our study offers insight into how model capacity shapes learning behavior and offers broader implications for the design and deployment of small language models.
\end{abstract}

\section{Introduction}
Large language models (LLMs) have shown impressive capabilities in both factual recall and pattern-based reasoning \cite{hurst2024gpt, team2023gemini}. However, in classical machine learning, there is a long-standing tradeoff between memorization and generalization, where improving one often comes at the cost of the other. We hypothesize that this same trade‐off governs LLM pre‐training. We propose that limited-capacity models are biased toward learning underlying rules, whereas larger models tend to memorize training examples at the expense of extrapolation.

To explore this hypothesis, we draw on the classical machine learning view of regularization, where models with limited capacity are naturally encouraged to generalize. Our experiments are based on nanoGPT \cite{karpathy2024nanogpt}, a lightweight implementation of the GPT architecture \cite{radford2019language} that serves as the starting point for our study. We introduce targeted modifications to the codebase that support a broader range of model sizes and configurations, allowing us to systematically evaluate models across a range of capacity constraints.

We train a sequence of Transformer models \cite{vaswani2017attention} with increasing size, ranging from minimal single-layer networks to a multi-layer GPT-style model. All experiments are conducted at the character level, ensuring that generalization and memorization arise from raw sequence modeling rather than token-level abstractions. These models are evaluated on two synthetic datasets specifically designed to isolate generalization and memorization. The first dataset assesses algorithmic generalization by holding out specific arithmetic expressions during training. The second dataset tests pure memorization using a small set of factual statements about capital cities. Together, these tasks allow us to isolate and analyze how model capacity influences learning behavior.

Our results show a consistent pattern: smaller models are able to learn and extrapolate the underlying arithmetic rules but fail to memorize factual content, while larger models excel at factual recall but struggle with extrapolation. Intermediate models display a potential shift between these behaviors. Across all configurations, no model achieves both generalization and memorization simultaneously.

These findings suggest that the trade-off between memorization and generalization is an inherent consequence of model scale and task composition, rather than a byproduct of overfitting or optimization. Our contributions include: (1) a controlled setup that cleanly isolates generalization and memorization, (2) an empirical analysis showing how increasing capacity shifts learning from rule-based generalization to pattern memorization, and (3) evidence that simultaneous exposure to both task types suppresses generalization across all model sizes.

\section{Background}

The trade-off between memorization and generalization in LLMs remains an open question. Nezhurina et al. \cite{nezhurina2025alicewonderlandsimpletasks} demonstrate that even state-of-the-art models (e.g., GPT-4 \cite{achiam2023gpt}, Claude 3 Opus) can fail catastrophically on simple reasoning tasks explicitly designed to test generalization. These failures raise concerns about the robustness of LLM reasoning and suggest a potential tension between generalization and pattern memorization. In contrast, Tirumala et al. \cite{tirumala2022memorizationoverfittinganalyzingtraining} show that large models can memorize substantial portions of their training data without exhibiting classical signs of overfitting, suggesting that memorization does not always interfere with generalization. This apparent contradiction highlights the need for controlled experiments that isolate these behaviors during model training.

During pre-training, LLMs absorb vast and diverse corpora to build broad linguistic competence and world knowledge, but this process can conflate rote memorization with deeper generalization. Wang et al.~\cite{wang2025generalizationvsmemorizationtracing} show that knowledge‑intensive tasks benefit from distributional memorization, while reasoning‑intensive tasks benefit from distributional generalization. Maltoni et al.~\cite{Maltoni_2024} show that nanoGPT, when pre-trained on binary‑arithmetic data, generalizes to 7‑bit addition and multiplication rather than merely memorizing examples. Together, these results suggest that systematic generalization in LLMs depends not only on the scale of pre-training, but also critically on the nature of the training distribution and the way tasks are represented.

After pre-training, LLMs are typically adapted via supervised fine-tuning (SFT) \cite{pareja2024unveilingsecretrecipeguide} on labeled examples. While effective at aligning model outputs, SFT can reinforce patterns in the fine-tuning data and amplify memorization. Chu et al.\cite{chu2025sftmemorizesrlgeneralizes} find that SFT models often replicate training examples verbatim, especially on narrow or repetitive datasets. Zeng et al.\cite{zeng2024exploringmemorizationfinetunedlanguage} further show that memorization increases with longer training and smaller datasets. These behaviors raise concerns about the robustness and generalization of SFT-aligned models.

The final stage of LLM training is reinforcement learning. Reinforcement learning with human feedback (RLHF)~\cite{ouyang2022traininglanguagemodelsfollow} improves generalization via preference alignment, but can reduce output diversity~\cite{kirk2024understandingeffectsrlhfllm} and often reverses memorization from SFT~\cite{chu2025sftmemorizesrlgeneralizes}. Recently, reinforcement learning with verifiable rewards (RLVR) has emerged as a scalable alternative. Wang et al.~\cite{wang2025reinforcementlearningreasoninglarge} show that a single verifiable example can dramatically improve math reasoning. 

To investigate LLM dynamics in a controlled setting, we take a bottom up approach focused on model pre-training.

\section{Methods}

\subsection{Datasets}
We design two fully synthetic, character-level benchmarks to isolate generalization versus memorization:

\textbf{Arithmetic:} This dataset consists of input–output pairs representing simple arithmetic expressions for addition and subtraction: \texttt{\textless a+b=c\textgreater} and \texttt{\textless a-b=c\textgreater}, where $a, b \in \{0, \dots, 9\}$ ; "\textless" and "\textgreater" are start and stop characters respectively. To test for true algorithmic extrapolation, all expressions involving both 5 and 7 (i.e., \texttt{5+7}, \texttt{7+5}, \texttt{5-7}, \texttt{7-5}) are withheld from training and validation.

Each line in the dataset is 9 characters long; shorter expressions are padded accordingly. We use a block size of 9 to provide full context per math sequence and a batch size of 882. This corresponds to half of the total dataset (196 expressions $\times$ 9 characters = 1764), ensuring that each batch covers a representative subset of the data without passing the full dataset each iteration.

\textbf{Capital Cities:} This dataset contains 50 factual statements of the form ``Capital of X is Y,'' designed to probe pure memorization. The same 50 examples are used for both training and validation. The longest sequence is 24 characters, so we set block size to 24. The batch size is set to 576, a square number (24 $\times$ 24) that ensures evenly sized batches.

\textbf{Combined Test:} For evaluations that include both arithmetic and factual data, we use the same batch size of 576 and block size of 24 as the Capital Cities dataset. This ensures compatibility, as 576 is divisible by 9, allowing arithmetic examples to fit cleanly into the same batch.

We intentionally use the same data for both training and validation. This setup reflects the kind of data duplication often found in LLM pre-training. It also allows us to isolate memorization dynamics more effectively, while still taking advantage of built-in checkpointing based on validation loss.

\subsection{Model Architectures}

To systematically explore the effects of model capacity, we constructed a series of models distinguished by their embedding size ($n$). Each model is referred to by its $n$ value. For example, \textbf{n14} corresponds to $n = 14$.

To support smaller nanoGPT models, we introduce an MLP expansion parameter ($m$), which overrides the default MLP hidden layer size of $4n$ with a user-specified value $mn$, where $n$ is the embedding size and $m$ is the expansion factor.

We began with the smallest possible configuration: one layer, one attention head, $n = 1$, and $m = 1$. From this minimal setup, we incrementally increased $n$ until the model successfully learned a basic arithmetic task, identifying $n = 14$ as the smallest viable model (\textbf{n14}). We then doubled $n$ to create \textbf{n28} and \textbf{n56}.

In addition to these single-layer, single-head models, we also evaluated a larger GPT-style architecture referred to as the \textbf{Multi-Layer Transformer (MLT)}. This model uses multiple layers and attention heads, following the Shakespeare training configuration from the nanoGPT repository. Although not identical to GPT-2, it is representative of standard small-scale GPT-style models and serves as a useful comparison point.

\subsection{Implementation Details}

We extend the original method from nanoGPT \cite{karpathy2024nanogpt} by adding evaluations and further hyperparameter customization. This evaluation runs every 250 steps and reports accuracy on our (0-9) arithmetic dataset and the capital cities tasks. For arithmetic, this includes the \texttt{(5,7)} combinations that are excluded from the training and validation set. Each test input is evaluated 10 times to account for sampling variability, and the best combined evaluation score across training is logged for each model and displayed in the results tables. All training was conducted on a single NVIDIA RTX 5090 GPU.  Other than these changes, the training setup remains consistent with that of the original nanoGPT, using standard token-level cross-entropy loss.

To ensure reproducibility, we globally seed Python’s random, NumPy, and PyTorch (both CPU and all CUDA devices), and enforce deterministic behavior in cuDNN by disabling benchmarking. Other than these changes, the training setup remains consistent with that of the original nanoGPT, including the use of cross-entropy loss.

All models were trained with identical optimization settings: \texttt{max\_iters = 30000}, \texttt{lr\_decay\_iters = 25000}, and \texttt{beta2 = 0.99}, ensuring a fair comparison across scales. Each model was trained from scratch in every experiment with the hyperparameters found in Table \ref{tab:hyper}.

\begin{table}[ht]
\centering
\caption{Hyperparameter comparison: n-models vs.\ MLT}
\begin{tabular}{lcc}
\toprule
\textbf{Parameter}       & \textbf{n-models} & \textbf{MLT} \\
\midrule
Embedding size ($n$)     & 14, 28, 56        & 384 \\
Layers                   & 1                 & 6 \\
Attention heads          & 1                 & 6 \\
MLP expansion ($m$)      & 1                 & 4 \\
Dropout                  & 0.0               & 0.2 \\
Weight decay             & 0                 & 0.1 \\
Learning rate            & $1 \times 10^{-2}$ & $1 \times 10^{-5}$ \\
Min.\ learning rate      & $1 \times 10^{-4}$ & $1 \times 10^{-6}$ \\
\bottomrule
\end{tabular}
\label{tab:hyper}
\end{table}

To confirm that the observed scaling effects were not due to differences in regularization, we conducted a control experiment where all models were retrained with the same settings: weight decay set to 0.1 and dropout to 0.0. We refer to this setup as the \textit{Controlled Regularization} condition. These experiments used the same training protocol as the original runs and are presented alongside the main results. Tables and figures are labeled accordingly, and full details, including accuracy metrics and training curves, are provided in Appendix~\ref{sec:controlled_reg}.

\section{Results}
All findings reported below are consistent under Controlled Regularization (see Appendix~\ref{sec:controlled_reg}).

\subsection{Extrapolation on Held-Out Arithmetic Cases}

\begin{table}[ht]
  \centering
  \begin{threeparttable}
      \caption{Addition \& Subtraction Performance}
      \label{tab:math}
      \begin{tabular}{l c c c c}
        \toprule
        Model & Parameters & Addition & Subtraction & (5,7)\tnote{*} \\
        \midrule
        n14   &  1.46k  & 100.0\% & 100.0\% & 40/40\ \\
        n28   &  5.26k  &  98\% &  98\% & 0/40 \\
        n56   & 19.94k  &  98\% &  98\% & 0/40 \\
        MLT   & 10.63M  &  98\% &  98\% & 0/40 \\
        \bottomrule
      \end{tabular}
    \begin{tablenotes}
      \item[*] The four combinations of (5,7) 10 attempts each
    \end{tablenotes}
  \end{threeparttable}
\end{table}

\begin{figure}[!ht]
  \centering
  \includegraphics[width=\columnwidth]{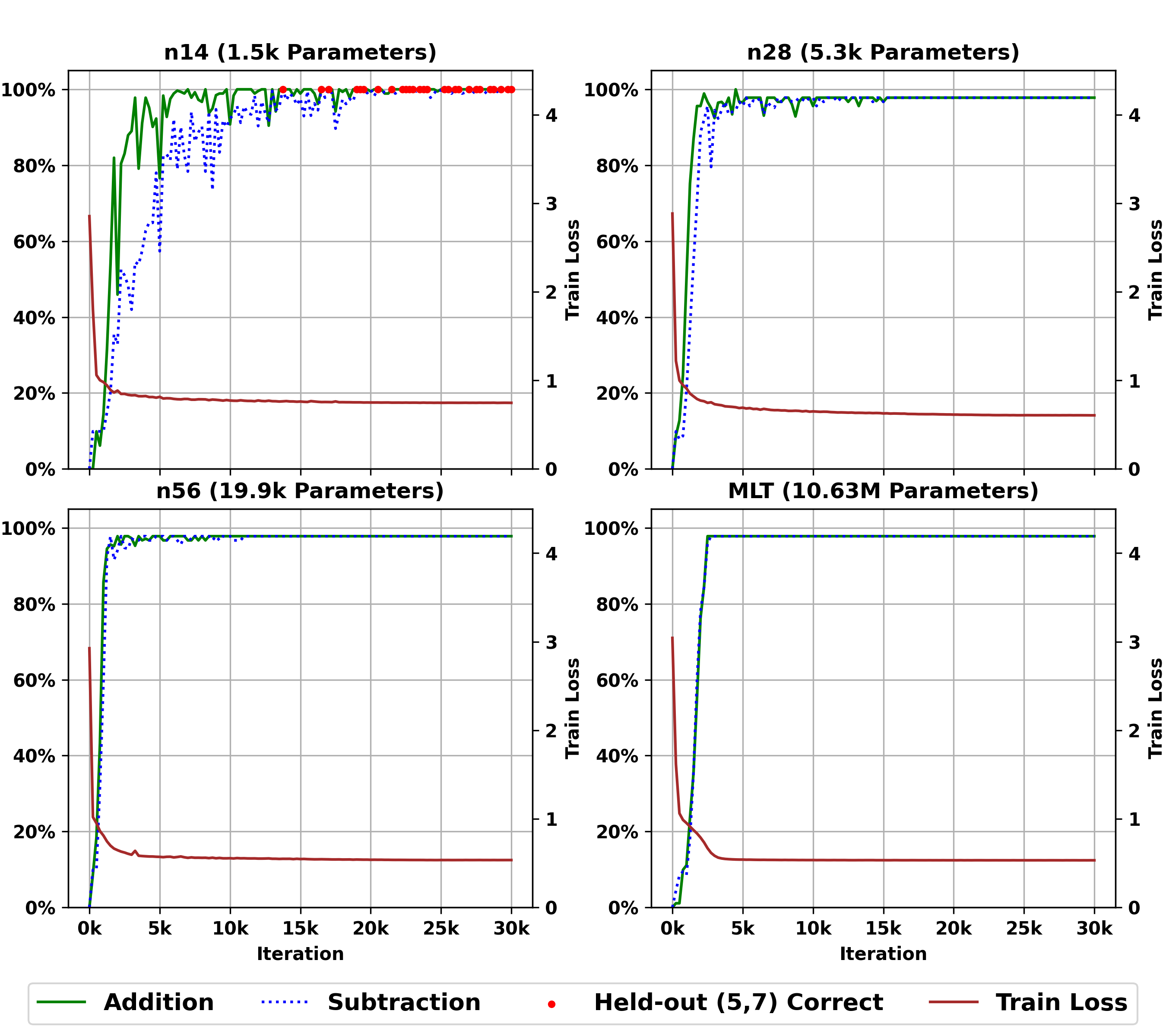}
  \caption{Addition and subtraction accuracy over the math evaluation, which includes the four withheld $(5,7)/(7,5)$ cases. Red dots indicate where both curves hit 100\%, signifying genuine extrapolation to the unseen pairs. Only n14 ever earns these red markers; larger models plateau below perfect overall accuracy due to being unable to extrapolate to the withheld combinations.}
  \label{fig:grid_math}
\end{figure}

The results in Table~\ref{tab:math} reveal a clear distinction between memorization and generalization. The smallest model, n14, achieves perfect accuracy on both addition and subtraction tasks, including all 40 held out (5,7) cases, showing limited but effective algorithmic generalization. In contrast, all larger models (n28, n56, MLT) perform well on seen expressions but fail entirely on the held out (5,7) inputs.

Figure~\ref{fig:grid_math} shows that n14 reaches perfect (100\%) accuracy multiple times during training, indicating successful extrapolation to the withheld $(5,7)$ combinations. Interestingly, the n28 model briefly predicted a couple (5,7) addition cases correctly in early training, but this behavior disappeared after 5k iterations and never emerged for subtraction. This suggests a possible transition from generalization to memorization as capacity increases.

The Arithmetic results are consistent with what we observe under the Controlled Regularization condition (Appendix~\ref{sec:math_wd}). In contrast, the larger models (n28, n56, MLT) plateau at 98\%, failing on every withheld case while performing flawlessly on seen inputs. These results support the case for decreasing Transformer model capacity to encourage generalization over memorization.

\subsection{Factual Recall of Capital Cities}

\begin{table}[ht]
  \centering
  \caption{Memorization of Capital Cities Performance}
  \label{tab:facts}
  \begin{tabular}{l c c}
    \toprule
    Model & Parameters & Facts Accuracy \\
    \midrule
    n14   &  1.93k  &   8.2\% \\
    n28   &  6.22k  & 100.0\% \\
    n56   & 21.84k  & 100.0\% \\
    MLT   & 10.64M  & 100.0\% \\
    \bottomrule
  \end{tabular}
\end{table}

\begin{figure}[!ht]
  \centering
  \includegraphics[width=\columnwidth]{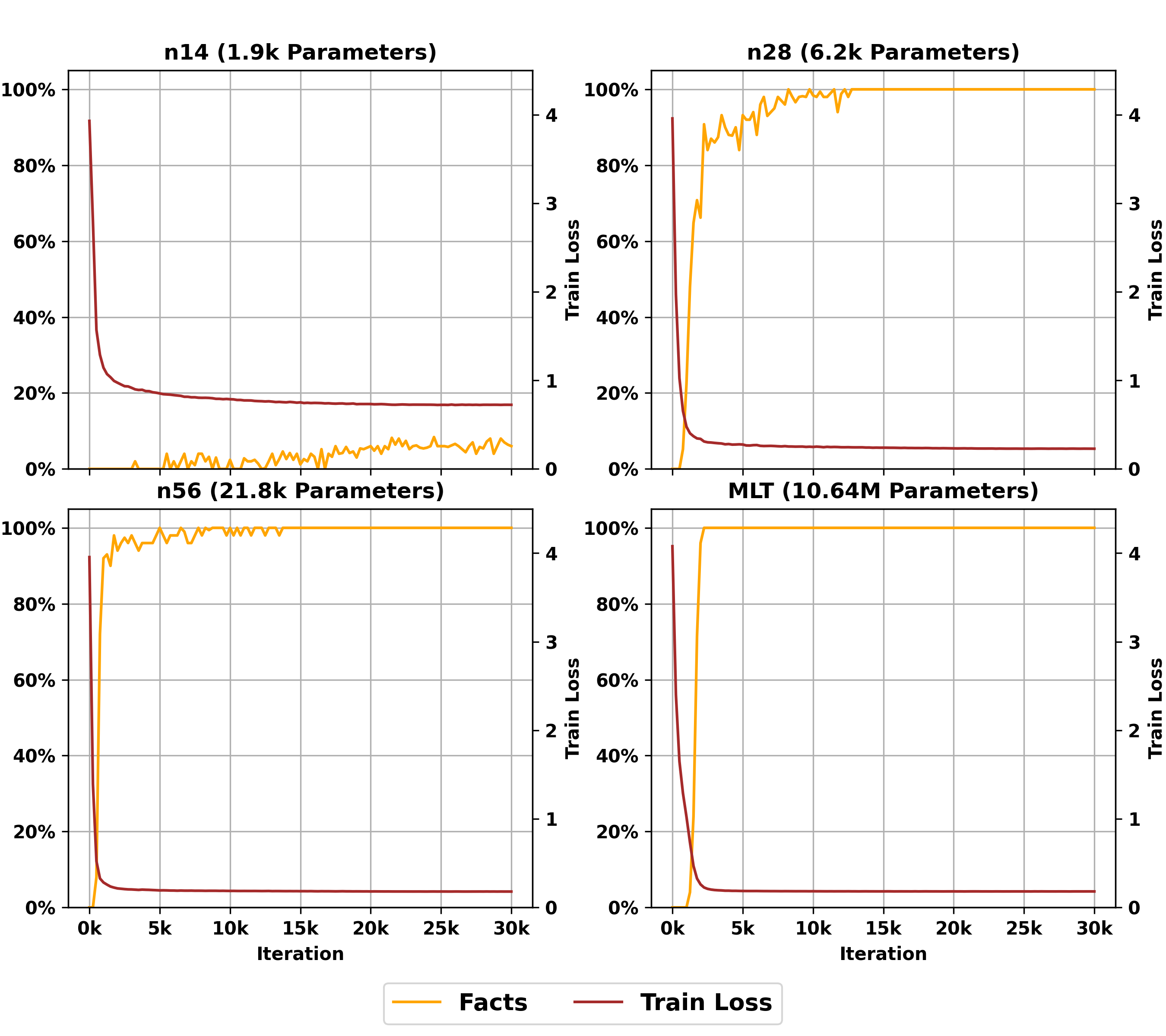}
  \caption{Training Dynamics on Factual Recall: Accuracy over 30k iterations for factual memorization task (Capital Cities). Only models at or above the n28 scale achieve full memorization; the n14 model fails to converge. This illustrates a clear capacity threshold required for factual recall.}  \label{fig:grid_facts}
\end{figure}

Table~\ref{tab:facts} demonstrates that memorization requires a minimum threshold of model capacity. The smallest model, n14, achieves only 8.2\% accuracy, failing to retain the majority of factual statements. However, a increase in parameter count yields dramatic improvements, with n28, n56 and MLT achieving perfect recitation of the capital cities.

In Figure~\ref{fig:grid_facts}, we show how factual memorization progresses over time for each model. The n14 model fails to converge throughout training. In contrast, n28 and larger models quickly achieve perfect accuracy, with minimal variance. These curves emphasize how memorization appears only after a clear capacity threshold is crossed, underscoring that memorization, unlike generalization, requires sufficient parameter count to store direct mappings.

This capacity threshold for factual recall remains stable for the Controlled Regularization experiemnt (Appendix~\ref{sec:facts_wd}), reaffirming that the result is intrinsic to model size, not due to architectural regularization choices.

\subsection{Joint Arithmetic and Capital Cities Training}

\begin{table*}[!t] 
  \centering
  \begin{threeparttable}
  \caption{Combined Arithmetic and Capital Cities Learning Performance}
  \label{tab:combined}
  \begin{tabular}{l c c c c c c}
    \toprule
    Model & Parameters & Addition & Subtraction & Facts & Combined & (5,7)\tnote{*} \\
    \midrule
    n14   &  2.14k  & 31.2\% & 39.4\%  &  2.0\%  & 28.6\% & 0/40 \\
    n28   &  6.64k  & 95.0\% & 97\%  & 87.6\%  & 94.3\% & 0/40 \\
    n56   & 22.68k  & 98\% & 98\%  & 100.0\% & 98.4\% & 0/40\\
    MLT   & 10.65M  & 98\% & 98\%  & 100.0\% & 98.4\% & 0/40 \\
    \bottomrule
  \end{tabular}
  \begin{tablenotes}
      \item[*] The four combinations of (5,7) 10 attempts each
    \end{tablenotes}
  \end{threeparttable}
\end{table*}

When arithmetic and factual tasks are trained jointly (Table~\ref{tab:combined}), none of the tested models correctly evaluate the held-out (5,7) arithmetic cases. However, relative to the isolated training experiments, the joint setting reveals more pronounced differences in model capacity. The n14 model shows a sharp decline, scoring below 40\% on addition and subtraction, despite being the only model that previously extrapolated successfully in the arithmetic only setting. Its accuracy on the capital cities task also falls from 8.2\% to just 2\%. These results suggest that the introduction of the capitals task distracts from learning the underlying arithmetic rule, while overwhelming the model's limited capacity.

The intermediate-capacity n28 model exhibits a similar pattern of capacity saturation when trained on both tasks. While it previously achieved 98\% accuracy on both addition and subtraction (with the remaining 2\% corresponding to the held-out (5,7) cases in both modalities), it now hovers around 95\%. Additionally, its factual accuracy on the capital cities task decreases from 100\% to approximately 88\%. The model appears to lack sufficient capacity to memorize the training data as effectively as it did in the isolated tasks. This limitation should impose implicit pressure on the model to learn the underlying rule. However, this pressure fails to produce successful extrapolation to the withheld (5,7) cases.

The larger n56 and MLT models retain strong performance on in-distribution examples from both tasks. However, they also fail to extrapolate to the held-out arithmetic cases. This pattern implies that when exposed to competing learning objectives, increased capacity primarily enhances the ability to memorize additional domains, rather than promoting the abstraction of general rules.

\begin{figure}[!ht]
  \centering
  \includegraphics[width=\columnwidth]{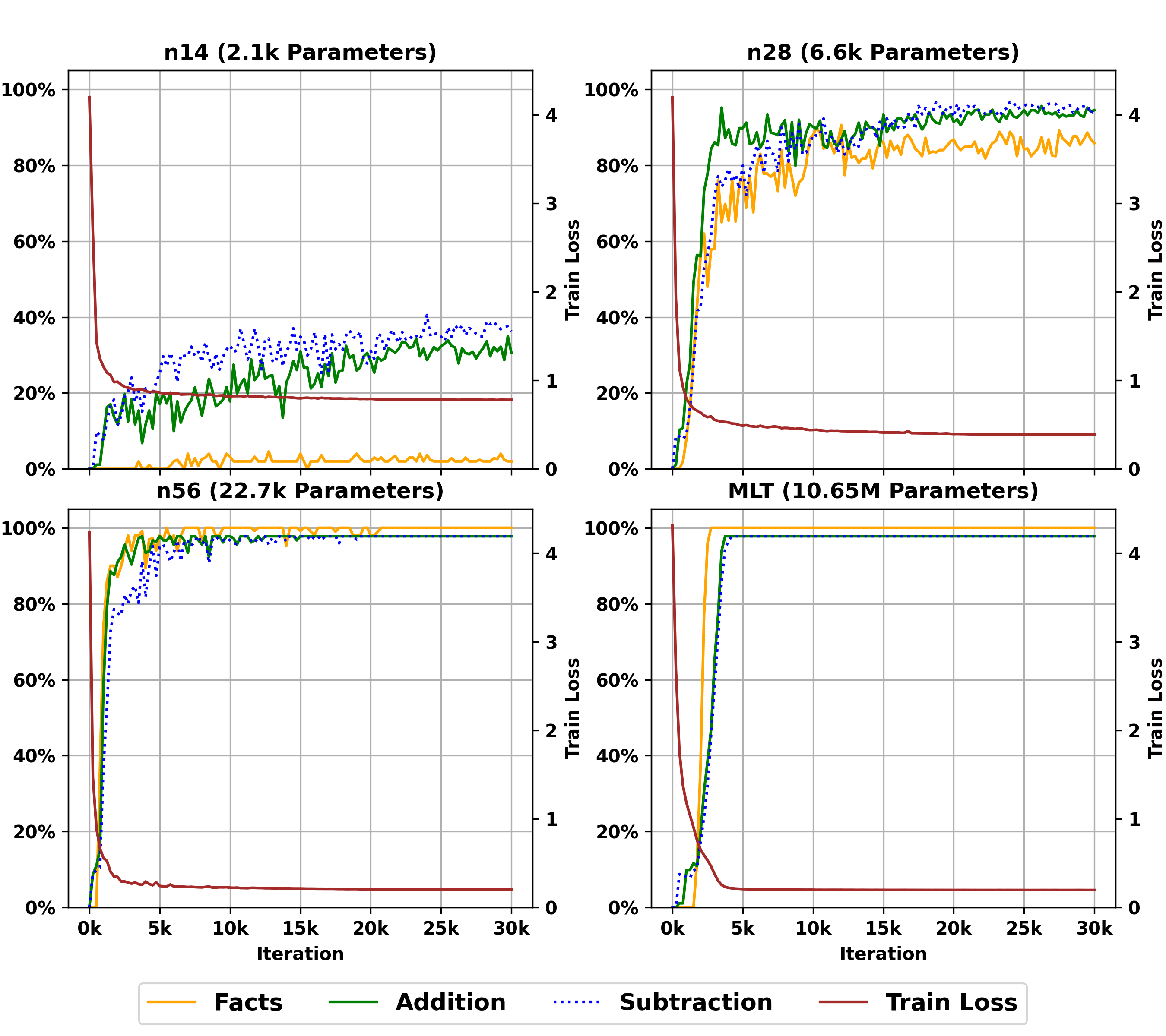}
  \caption{Training Dynamics on Joint Task (Arithmetic + Factual): Accuracy over 30k iterations for models trained on both tasks. No model generalizes to held-out (5,7) arithmetic cases. The n14 model regresses relative to its arithmetic-only performance, and all larger models prioritize memorization. This reveals the incompatibility of generalization and memorization under simultaneous training conditions.}  \label{fig:grid_combined}
\end{figure}

Plotting the accuracy of the combined tasks over iterations in Figure~\ref{fig:grid_combined} illustrates the training trajectories. No model is able to generalize to the held-out arithmetic combinations even when evaluated every 250 iterations. This suggests that the introduction of a second task disrupts the model’s ability to learn and apply abstract rules, regardless of model size.

Controlled Regularization did not alter this trend (Appendix~\ref{sec:combined_wd}). Even when regularization settings were held constant, the tension between memorization and extrapolation persisted, underscoring a deeper incompatibility between the two learning modes under joint training.

To ensure the large MLT model had ample opportunity to optimize its much larger set of parameters, we conducted an extended run of 1.5 million iterations (Appendix~\ref{sec:grokking}). Even under this prolonged regime, the model failed to generalize, reinforcing the interpretation that memorization represents a stable solution not easily escaped, rather than a transient phase preceding generalization.

\section{Discussion}

Our findings reveal a fundamental tension between memorization and generalization in Transformer-based models. This holds across both the original and Controlled Regularization settings (Appendix~\ref{sec:controlled_reg}), suggesting it is not merely a consequence of regularization choices. Analysis over the course of training shows that subtraction is consistently harder for the model to learn than addition, likely due to its non-commutative nature. These patterns indicate that both model capacity and task structure shape learning outcomes: given an identical task, smaller models are pressured to extract underlying structure, whereas larger models tend to converge quickly to a stable memorization of the training data.

\subsection{Extrapolation vs. Generalization}

To test for generalization beyond the training domain, we evaluated the pre-trained n14 model on arithmetic problems with inputs ranging from 0 to 19. After excluding equations that involved only single-digit numbers (0 to 9), we assessed the model’s performance on novel cases. The results were clear: 0\% accuracy on addition and just 4.75\% on subtraction. This sharp decline suggests that the model did not learn a general arithmetic procedure. The slightly better performance on subtraction may reflect the difficulty the model faced in learning it, which may have encouraged a small degree of generalization. Nonetheless, since n14 solves unseen single-digit problems like \texttt{5+7} and \texttt{7-5} perfectly but fails on even slightly out-of-distribution cases, we interpret this not as evidence of generalization, but as narrow extrapolation within the training domain.

\subsection{Trade-Off Between Memorization and Generalization}

The broader pattern observed in our results aligned with our central hypothesis: smaller models, constrained by limited capacity, cannot memorize training data verbatim and are thus compelled to learn and apply underlying structure. Larger models, in contrast, have sufficient capacity to memorize input-output mappings and tend to rely on pattern matching rather than rule extraction.

This trade-off appears rooted in the model’s \textit{representational limits}. From a learning-theoretic perspective, capacity-constrained models exhibit an implicit \textit{inductive bias}: lacking the ability to store all observed data, they are biased toward simpler, more compressible hypotheses such as structural patterns that generalize. This reflects the classical notion of a \textit{simplicity prior}, wherein models favor low-description-length solutions \cite{friedland2024information}. In our setting, this manifests as rule-based extrapolation in arithmetic tasks. As capacity increases, however, this inductive bias weakens, and models tend to rely on high-variance memorization rather than internalizing structural generalizations.

For small language models (SLMs), this insight is particularly salient. Though often dismissed as underpowered, our results suggest that their limitations may actively promote abstraction and generalization in reasoning tasks. Conversely, SLMs may struggle with tasks requiring fine-grained memorization.

\subsection{Implications for Hallucination Mitigation}

One of the most consequential implications of our work concerns hallucination. Optimizing aggressively for factual recall may come at the cost of generalization. When our smallest model (n14) is jointly trained on both arithmetic and factual data, it fails to learn either, and fails to extrapolate to (5,7) combinations even though it previously succeeded under isolated conditions. This suggests that pressure to memorize suppresses the emergence of generalizable rules.

This implies that hallucination may in part stem from deeper architectural and capacity-based trade-offs: attempting to encode too much factual specificity within a model may actively impair its ability to generalize. This highlights the need to reconsider prevailing training paradigms, especially when generalization is critical for safe and robust deployment.

\section{Limitations}

While the n14 model appears to generalize to held-out arithmetic expressions, its performance collapses on multi-digit inputs. This indicates that its learned rules apply only narrowly within the 0--9 digit space. As such, we interpret this behavior as shallow extrapolation rather than robust generalization. A deeper investigation of its behavior on more complex arithmetic tasks lies outside the present scope and is left to future work.

\section{Conclusion}

This work investigates the trade-off between memorization and generalization in Transformer-based language models, demonstrating that this tension is not simply a byproduct of training dynamics but a structural consequence of model capacity. By pre-training models from scratch on controlled, synthetic tasks explicitly designed to support generalization, we find that small models exhibit only limited extrapolation to unseen arithmetic cases. In contrast, larger models show no tendency toward extrapolation or structural generalization, instead defaulting to memorization of observed patterns. Intermediate-capacity models hint at the transition between these behaviors, reinforcing the capacity-dependent nature of this trade-off.

Critically, none of the model sizes tested were able to reconcile both behaviors when jointly trained on reasoning and factual recall tasks. These findings highlight an inherent limitation of monolithic model architectures: the representational capacity that enables memorization may actively suppress generalization. For small language models, this constraint poses both a challenge and an opportunity. The limited capacity may hinder factual recall but promote abstraction and structural reasoning, particularly in domains like robotics or embedded systems, where generalization is crucial and compute is limited.

Our results suggest that future language model design should explicitly account for this trade-off, particularly in the context of pre-training objectives. Modular or specialized architectures that separate memorization from generalization pathways may offer a promising path forward. Future work will explore whether these dynamics extend to real-world domains and how training strategies can be adapted to better balance the two learning modes.

\section*{Acknowledgments}

We would like to acknowledge Dr. Gerald Friedland for inspiring the investigation into the memorization in transformers through his class and book \cite{friedland2024information}. We would also like to thank MD Sunbeam for his insight and for reviewing the initial draft of the paper.

\section*{Impact Statement}

This paper presents work whose goal is to advance the field of 
Machine Learning. There are many potential societal consequences 
of our work, none which we feel must be specifically highlighted here.

\nocite{langley00}

\bibliography{references}

\begin{thebibliography}{20}
\providecommand{\natexlab}[1]{#1}
\providecommand{\url}[1]{\texttt{#1}}
\expandafter\ifx\csname urlstyle\endcsname\relax
  \providecommand{\doi}[1]{doi: #1}\else
  \providecommand{\doi}{doi: \begingroup \urlstyle{rm}\Url}\fi

\bibitem[Achiam et~al.(2023)Achiam, Adler, Agarwal, Ahmad, Akkaya, Aleman, Almeida, Altenschmidt, Altman, Anadkat, et~al.]{achiam2023gpt}
Achiam, J., Adler, S., Agarwal, S., Ahmad, L., Akkaya, I., Aleman, F.~L., Almeida, D., Altenschmidt, J., Altman, S., Anadkat, S., et~al.
\newblock Gpt-4 technical report.
\newblock \emph{arXiv preprint arXiv:2303.08774}, 2023.

\bibitem[Chu et~al.(2025)Chu, Zhai, Yang, Tong, Xie, Schuurmans, Le, Levine, and Ma]{chu2025sftmemorizesrlgeneralizes}
Chu, T., Zhai, Y., Yang, J., Tong, S., Xie, S., Schuurmans, D., Le, Q.~V., Levine, S., and Ma, Y.
\newblock Sft memorizes, rl generalizes: A comparative study of foundation model post-training, 2025.
\newblock URL \url{https://arxiv.org/abs/2501.17161}.

\bibitem[Friedland(2024)]{friedland2024information}
Friedland, G.
\newblock \emph{Information-Driven Machine Learning: Data Science as an Engineering Discipline}.
\newblock Springer, 1st edition, 2024.
\newblock 1st ed. 2024 Edition.

\bibitem[Hurst et~al.(2024)Hurst, Lerer, Goucher, Perelman, Ramesh, Clark, Ostrow, Welihinda, Hayes, Radford, et~al.]{hurst2024gpt}
Hurst, A., Lerer, A., Goucher, A.~P., Perelman, A., Ramesh, A., Clark, A., Ostrow, A., Welihinda, A., Hayes, A., Radford, A., et~al.
\newblock Gpt-4o system card.
\newblock \emph{arXiv preprint arXiv:2410.21276}, 2024.

\bibitem[Karpathy(2024)]{karpathy2024nanogpt}
Karpathy, A.
\newblock {nanoGPT}: The simplest, fastest repository for training/finetuning medium-sized gpts.
\newblock \url{https://github.com/karpathy/nanoGPT/tree/master}, 2024.
\newblock GitHub repository; commit \texttt{93a43d9} (Dec.\ 9, 2024); accessed May 19, 2025.

\bibitem[Kirk et~al.(2024)Kirk, Mediratta, Nalmpantis, Luketina, Hambro, Grefenstette, and Raileanu]{kirk2024understandingeffectsrlhfllm}
Kirk, R., Mediratta, I., Nalmpantis, C., Luketina, J., Hambro, E., Grefenstette, E., and Raileanu, R.
\newblock Understanding the effects of rlhf on llm generalisation and diversity, 2024.
\newblock URL \url{https://arxiv.org/abs/2310.06452}.

\bibitem[Liu et~al.(2022)Liu, Kitouni, Nolte, Michaud, Tegmark, and Williams]{liu2022understandinggrokkingeffectivetheory}
Liu, Z., Kitouni, O., Nolte, N., Michaud, E.~J., Tegmark, M., and Williams, M.
\newblock Towards understanding grokking: An effective theory of representation learning, 2022.
\newblock URL \url{https://arxiv.org/abs/2205.10343}.

\bibitem[Maltoni \& Ferrara(2024)Maltoni and Ferrara]{Maltoni_2024}
Maltoni, D. and Ferrara, M.
\newblock Arithmetic with language models: From memorization to computation.
\newblock \emph{Neural Networks}, 179:\penalty0 106550, November 2024.
\newblock ISSN 0893-6080.
\newblock \doi{10.1016/j.neunet.2024.106550}.
\newblock URL \url{http://dx.doi.org/10.1016/j.neunet.2024.106550}.

\bibitem[Nezhurina et~al.(2025)Nezhurina, Cipolina-Kun, Cherti, and Jitsev]{nezhurina2025alicewonderlandsimpletasks}
Nezhurina, M., Cipolina-Kun, L., Cherti, M., and Jitsev, J.
\newblock Alice in wonderland: Simple tasks showing complete reasoning breakdown in state-of-the-art large language models, 2025.
\newblock URL \url{https://arxiv.org/abs/2406.02061}.

\bibitem[Ouyang et~al.(2022)Ouyang, Wu, Jiang, Almeida, Wainwright, Mishkin, Zhang, Agarwal, Slama, Ray, Schulman, Hilton, Kelton, Miller, Simens, Askell, Welinder, Christiano, Leike, and Lowe]{ouyang2022traininglanguagemodelsfollow}
Ouyang, L., Wu, J., Jiang, X., Almeida, D., Wainwright, C.~L., Mishkin, P., Zhang, C., Agarwal, S., Slama, K., Ray, A., Schulman, J., Hilton, J., Kelton, F., Miller, L., Simens, M., Askell, A., Welinder, P., Christiano, P., Leike, J., and Lowe, R.
\newblock Training language models to follow instructions with human feedback, 2022.
\newblock URL \url{https://arxiv.org/abs/2203.02155}.

\bibitem[Pareja et~al.(2024)Pareja, Nayak, Wang, Killamsetty, Sudalairaj, Zhao, Han, Bhandwaldar, Xu, Xu, Han, Inglis, and Srivastava]{pareja2024unveilingsecretrecipeguide}
Pareja, A., Nayak, N.~S., Wang, H., Killamsetty, K., Sudalairaj, S., Zhao, W., Han, S., Bhandwaldar, A., Xu, G., Xu, K., Han, L., Inglis, L., and Srivastava, A.
\newblock Unveiling the secret recipe: A guide for supervised fine-tuning small llms, 2024.
\newblock URL \url{https://arxiv.org/abs/2412.13337}.

\bibitem[Power et~al.(2022)Power, Burda, Edwards, Babuschkin, and Misra]{power2022grokkinggeneralizationoverfittingsmall}
Power, A., Burda, Y., Edwards, H., Babuschkin, I., and Misra, V.
\newblock Grokking: Generalization beyond overfitting on small algorithmic datasets, 2022.
\newblock URL \url{https://arxiv.org/abs/2201.02177}.

\bibitem[Radford et~al.(2019)Radford, Wu, Child, Luan, Amodei, Sutskever, et~al.]{radford2019language}
Radford, A., Wu, J., Child, R., Luan, D., Amodei, D., Sutskever, I., et~al.
\newblock Language models are unsupervised multitask learners.
\newblock \emph{OpenAI blog}, 1\penalty0 (8):\penalty0 9, 2019.

\bibitem[Team et~al.(2023)Team, Anil, Borgeaud, Alayrac, Yu, Soricut, Schalkwyk, Dai, Hauth, Millican, et~al.]{team2023gemini}
Team, G., Anil, R., Borgeaud, S., Alayrac, J.-B., Yu, J., Soricut, R., Schalkwyk, J., Dai, A.~M., Hauth, A., Millican, K., et~al.
\newblock Gemini: a family of highly capable multimodal models.
\newblock \emph{arXiv preprint arXiv:2312.11805}, 2023.

\bibitem[Tirumala et~al.(2022)Tirumala, Markosyan, Zettlemoyer, and Aghajanyan]{tirumala2022memorizationoverfittinganalyzingtraining}
Tirumala, K., Markosyan, A.~H., Zettlemoyer, L., and Aghajanyan, A.
\newblock Memorization without overfitting: Analyzing the training dynamics of large language models, 2022.
\newblock URL \url{https://arxiv.org/abs/2205.10770}.

\bibitem[Vaswani et~al.(2017)Vaswani, Shazeer, Parmar, Uszkoreit, Jones, Gomez, Kaiser, and Polosukhin]{vaswani2017attention}
Vaswani, A., Shazeer, N., Parmar, N., Uszkoreit, J., Jones, L., Gomez, A.~N., Kaiser, {\L}., and Polosukhin, I.
\newblock Attention is all you need.
\newblock \emph{Advances in neural information processing systems}, 30, 2017.

\bibitem[Wang et~al.(2024)Wang, Yue, Su, and Sun]{wang2024grokkedtransformersimplicitreasoners}
Wang, B., Yue, X., Su, Y., and Sun, H.
\newblock Grokked transformers are implicit reasoners: A mechanistic journey to the edge of generalization, 2024.
\newblock URL \url{https://arxiv.org/abs/2405.15071}.

\bibitem[Wang et~al.(2025{\natexlab{a}})Wang, Antoniades, Elazar, Amayuelas, Albalak, Zhang, and Wang]{wang2025generalizationvsmemorizationtracing}
Wang, X., Antoniades, A., Elazar, Y., Amayuelas, A., Albalak, A., Zhang, K., and Wang, W.~Y.
\newblock Generalization v.s. memorization: Tracing language models' capabilities back to pretraining data, 2025{\natexlab{a}}.
\newblock URL \url{https://arxiv.org/abs/2407.14985}.

\bibitem[Wang et~al.(2025{\natexlab{b}})Wang, Yang, Zeng, Ren, Liu, Peng, Cheng, He, Wang, Gao, Chen, Wang, Du, and Shen]{wang2025reinforcementlearningreasoninglarge}
Wang, Y., Yang, Q., Zeng, Z., Ren, L., Liu, L., Peng, B., Cheng, H., He, X., Wang, K., Gao, J., Chen, W., Wang, S., Du, S.~S., and Shen, Y.
\newblock Reinforcement learning for reasoning in large language models with one training example, 2025{\natexlab{b}}.
\newblock URL \url{https://arxiv.org/abs/2504.20571}.

\bibitem[Zeng et~al.(2024)Zeng, Li, Ren, Liu, Xu, He, Xing, Wang, Tang, and Yin]{zeng2024exploringmemorizationfinetunedlanguage}
Zeng, S., Li, Y., Ren, J., Liu, Y., Xu, H., He, P., Xing, Y., Wang, S., Tang, J., and Yin, D.
\newblock Exploring memorization in fine-tuned language models, 2024.
\newblock URL \url{https://arxiv.org/abs/2310.06714}.

\end{thebibliography}
\bibliographystyle{icml2025}



\newpage
\onecolumn
\appendix
\section{Controlled Regularization Experiments}

To verify that the observed results were not artifacts of differing regularization settings, we re‑ran all experiments under “Controlled Regularization” (\texttt{dropout = 0.0}, \texttt{weight decay = 0.1}) with all other settings identical.
\label{sec:controlled_reg}

\subsection{Controlled Regularization: Arithmetic with (5,7) combinations withheld}
\label{sec:math_wd}

\begin{figure}[ht]
  \centering
  \includegraphics[width=0.8\columnwidth]{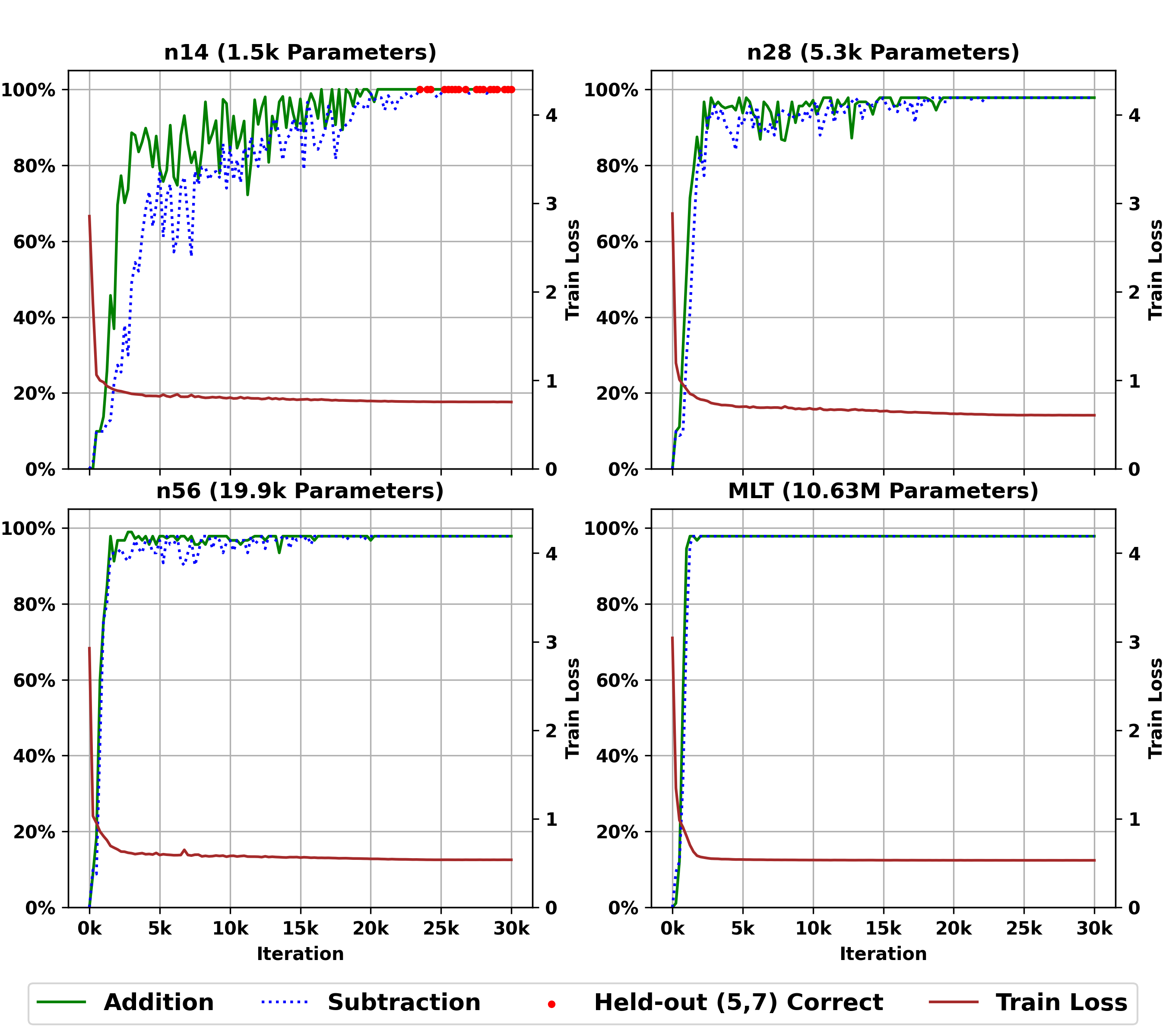}
  \caption{Arithmetic extrapolation with controlled regularization:
    only \textit{n14} generalizes to unseen (5,7); larger models plateau at 0\%.}
  \label{fig:appendix_math}
\end{figure}

These controlled‑regularization runs reproduce the main Arithmetic experiment’s pattern: only the n14 model extrapolates to the withheld (5,7) inputs, while all larger models plateau at zero (Table~\ref{tab:math_wd}, Fig.~\ref{fig:appendix_math}).

\begin{table}[ht]
  \centering
  \begin{threeparttable}
    \caption{Math Results (Controlled Regularization)}
    \label{tab:math_wd}
    \begin{tabular}{l c c c c}
      \toprule
      Model & Parameters & Addition & Subtraction & (5,7)\tnote{*} \\
      \midrule
      n14   &  1.46k  & 100.0\% & 100.0\% & 40/40 \\
      n28   &  5.26k  &  98\% &  98\% &  0/40 \\
      n56   & 19.94k  &  98\% &  98\% &  0/40 \\
      MLT   & 10.63M  &  98\% &  98\% &  0/40 \\
      \bottomrule
    \end{tabular}
    \begin{tablenotes}
      \item[*] The four withheld (5,7) combinations, 10 attempts each
    \end{tablenotes}
  \end{threeparttable}
\end{table}

\newpage
\subsection{Factual Recall}
\label{sec:facts_wd}

\begin{figure}[ht]
  \centering
  \includegraphics[width=0.8\columnwidth]{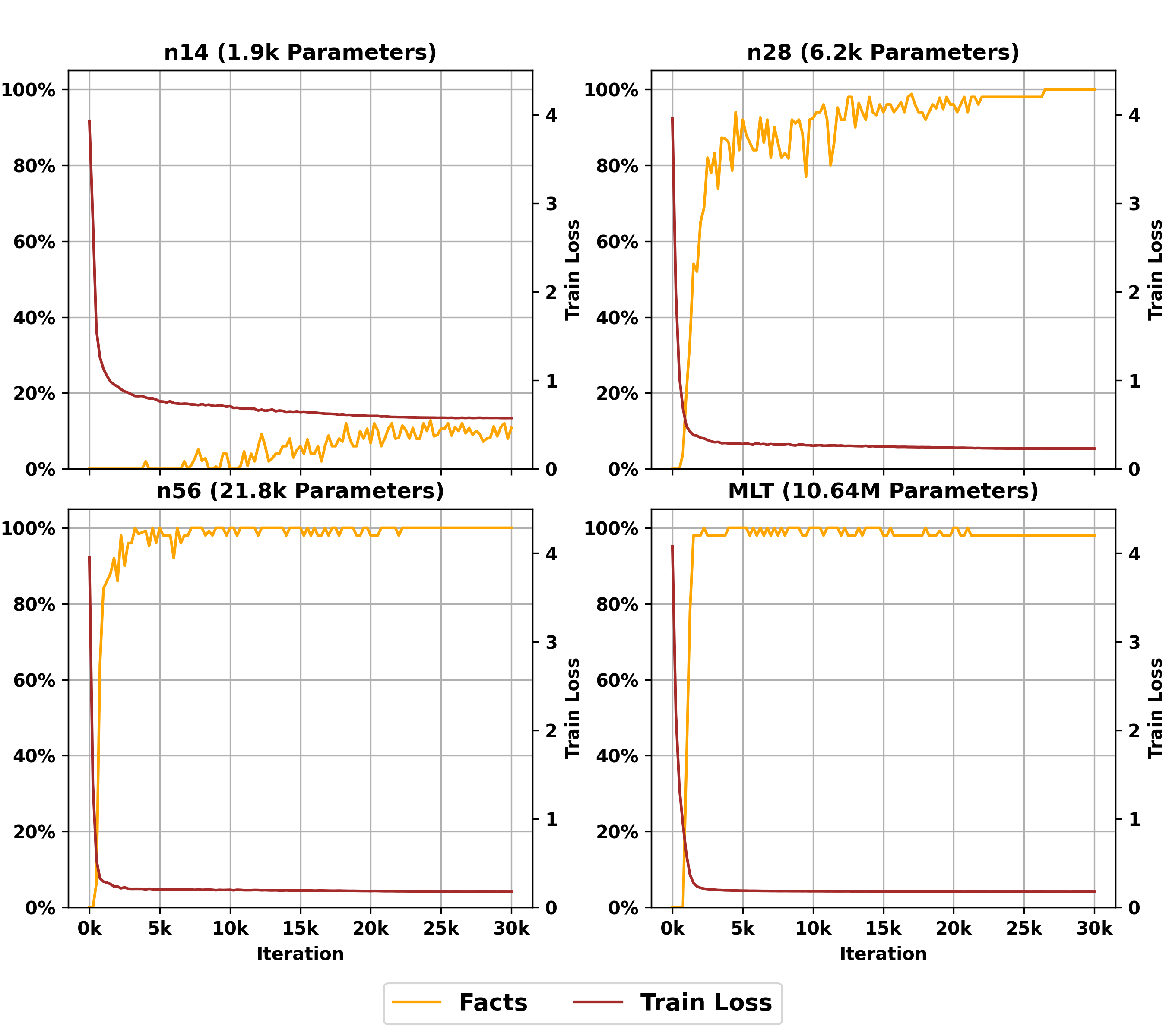}
  \caption{Factual recall under controlled regularization:
    \textit{n28} and above achieve perfect scores; \textit{n14} remains below.}
  \label{fig:appendix_facts}
\end{figure}

As in the main experiment, factual recall under controlled regularization shows a sharp capacity threshold: only models at n28 and above achieve perfect accuracy, while n14 continues to underfit (Table~\ref{tab:facts_wd}, Fig.~\ref{fig:appendix_facts}).

\begin{table}[ht]
  \centering
  \caption{Facts Results (Controlled Regularization)}
  \label{tab:facts_wd}
  \begin{tabular}{l c c}
    \toprule
    Model & Parameters & Facts Accuracy \\
    \midrule
    n14   &  1.93k  &  12.4\% \\
    n28   &  6.22k  & 100.0\% \\
    n56   & 21.84k  & 100.0\% \\
    MLT   & 10.64M  & 100.0\% \\
    \bottomrule
  \end{tabular}
\end{table}

\newpage
\subsection{Joint Task Performance}
\label{sec:combined_wd}

\begin{figure*}[ht]
  \centering
  \includegraphics[width=0.8\columnwidth]{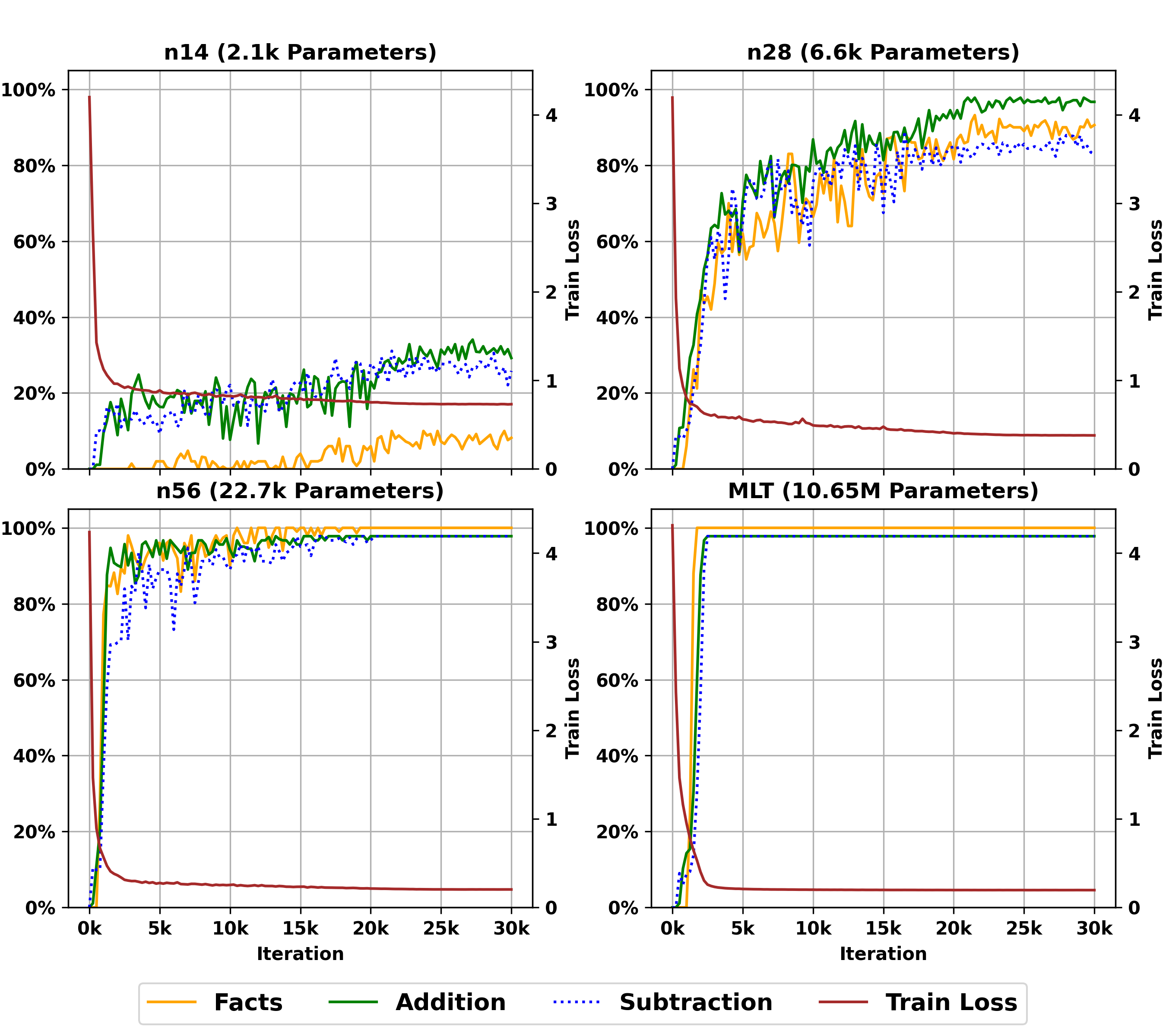}
  \caption{Joint performance with controlled regularization:
    no model extrapolates arithmetic; larger models still memorize facts.}
  \label{fig:appendix_combined}
\end{figure*}

Joint training again reveals the same tradeoff: no model extrapolates to (5,7) arithmetic, while larger models still memorize facts (Table~\ref{tab:combined_wd}, Fig.~\ref{fig:appendix_combined}). The interference pattern persists even under matched regularization.

\begin{table*}[ht]
  \centering
  \begin{threeparttable}
    \caption{Combined-Model Performance (Controlled Regularization)}
    \label{tab:combined_wd}
    \begin{tabular}{l c c c c c c}
      \toprule
      Model & Parameters & Addition & Subtraction & Facts & Combined & (5,7)\tnote{*} \\
      \midrule
      n14   &  2.14k   & 33.4\% & 28.5\% &  6.6\% & 26.1\% &  0/40 \\
      n28   &  6.64k   & 96.3\% & 88.7\% & 90.0\% & 92.0\% &  0/40 \\
      n56   & 22.68k  & 98\% & 98\%  & 100.0\% & 98.4\% & 0/40\\
      MLT   & 10.65M  & 98\% & 98\%  & 100.0\% & 98.4\% & 0/40 \\
      \bottomrule
    \end{tabular}
    \begin{tablenotes}
      \item[*] The four withheld (5,7) combinations, 10 attempts each
    \end{tablenotes}
  \end{threeparttable}
\end{table*}

\section{Grokking}
\label{sec:grokking}

Beyond the core training stages, we explored whether extended training might promote generalization, a phenomenon known as \textit{grokking}~\cite{power2022grokkinggeneralizationoverfittingsmall}. This dynamic, where models initially memorize the training set and only begin to generalize after prolonged training, has been observed in algorithmic tasks with small models. While intriguing, grokking is highly sensitive to model architecture, task formulation, and optimization details~\cite{wang2024grokkedtransformersimplicitreasoners, liu2022understandinggrokkingeffectivetheory}. Its occurrence in more complex or high-capacity models is not guaranteed, but was identified as an additional test to be ran. 

\subsection{Preliminary Grokking Check}

To investigate whether grokking might emerge under extended training, we conducted a one-off experiment using the arithmetic-only MLT task, extending training to 1.5 million iterations (roughly 24 hours of wall-clock time). We kept the model architecture and regularization fixed, modifying only the learning rate schedule (\(\mathrm{lr} = 1\times10^{-6}\), \(\mathrm{lr\_decay\_iters} = 1\times10^{7}\), \(\mathrm{min\_lr} = 1\times10^{-7}\)). Held-out evaluations were conducted every 250 iterations as in prior runs.

Despite this prolonged training, the model failed to produce any correct predictions on the 40 held-out (5,7) combinations. The lowest validation cross-entropy achieved was 0.5306, with no gain in generalization performance. This result suggests that, within our setup, memorization formed a stable attractor that was not overcome by time or training alone.

\begin{figure}[!ht]
  \centering
  \includegraphics[width=\columnwidth]{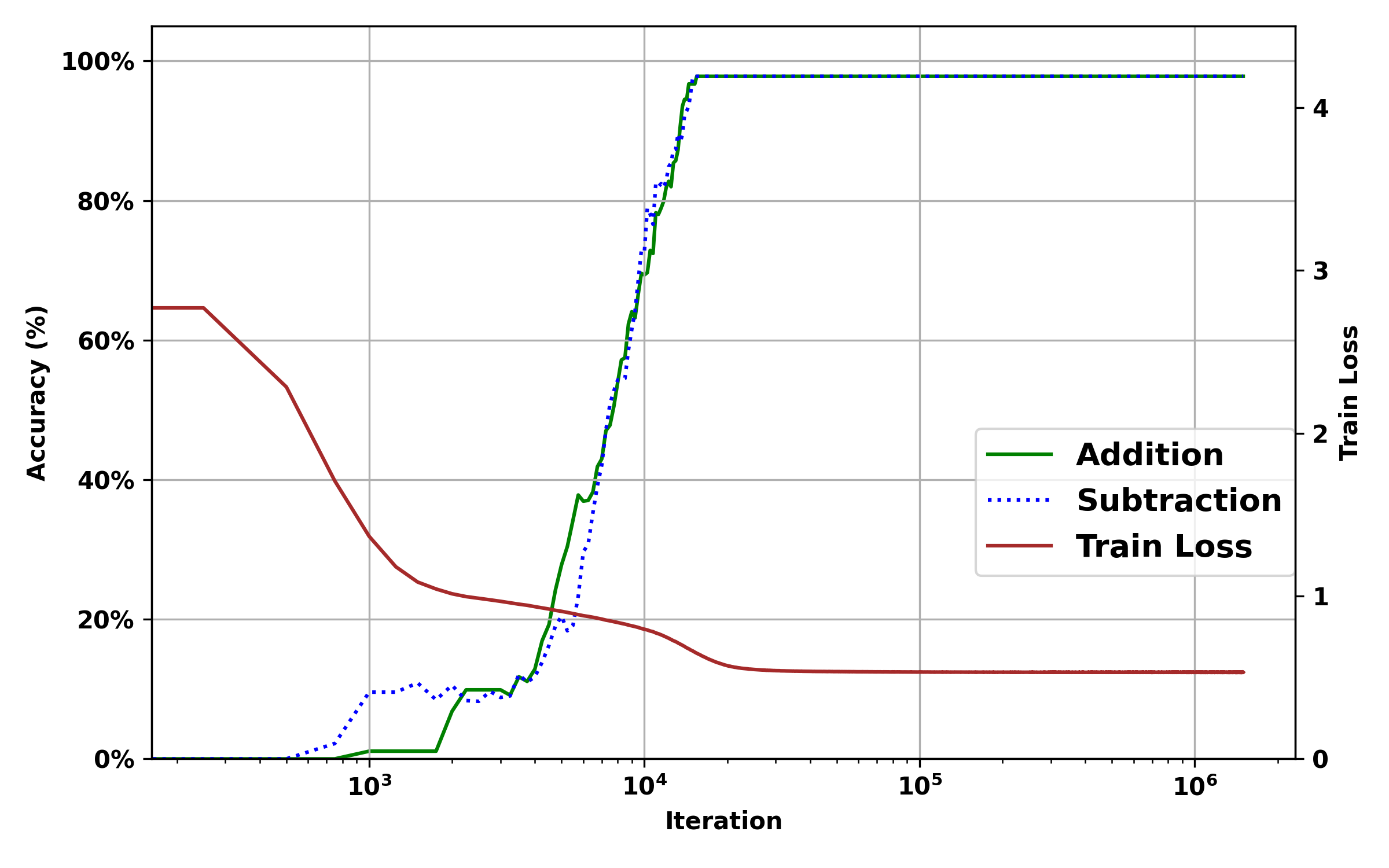}
  \caption{Extended training of the MLT model (10.63M parameters) on the arithmetic-only task for 1.5M iterations. X-axis is log-scaled. In-distribution addition and subtraction accuracies plateau near 98\%, while held-out (5,7) generalization remains at 0\%, indicating persistent memorization despite prolonged training.}
  \label{fig:grokking_attempt}
\end{figure}

Figure~\ref{fig:grokking_attempt} illustrates this: while in-distribution accuracy rapidly saturates and remains high, extrapolation performance stays flat at zero.

\subsection{When Grokking Fails to Emerge}

Although grokking has been reported in Transformer settings, particularly within a ``Goldilocks zone'' between memorization and confusion phases~\cite{liu2022understandinggrokkingeffectivetheory}, we did not observe this dynamic in our setting. Even after extensive training, the MLT model failed to generalize to unseen (5,7) examples, despite maintaining near-perfect accuracy on seen combinations.

This outcome suggests that in high-capacity models, memorization may constitute a persistent local optimum. While grokking points to a possible route toward generalization through structured representation learning, our results emphasize that this behavior is not guaranteed and may depend critically on model scale, task complexity, and the training regime.




\end{document}